\documentclass{article}
\usepackage{spconf,amsmath,graphicx,hyperref}
\usepackage{booktabs}
\usepackage{float}

\usepackage[T1]{fontenc}
\usepackage{lmodern}
\usepackage{microtype}
\usepackage[margin=1in]{geometry}

\usepackage{amsmath}
\usepackage{amssymb}
\usepackage{mathtools}
\usepackage{bm}

\usepackage{booktabs}
\usepackage{enumitem}

\title{Cross-Domain Joint-Embedding Pretraining for Heterogeneous Industrial Time Series}
\name{Jonathan Adam Rico$^{\star \dagger}$\thanks{This research was supported by
the SUTD Research Surplus Grant No. RS-INSUR-00019-E0601-S00.} \qquad Nagarajan Raghavan$^{\star}$ \qquad Senthilnath Jayavelu$^{\dagger}$}
  
  \address{$^{\star}$Engineering Product Development (EPD), SUTD, Singapore \\
      $^{\dagger}$Institute of Advanced Intelligence and Computing (IAIC), A*STAR, Singapore}

\begin{document}
%
\maketitle
%
\begin{abstract}
Industrial predictive maintenance and manufacturing tasks frequently contain abundant sensor measurements, but with limited labeled data. We investigate cross-domain self-supervised pretraining using joint-embedding predictive architecture for heterogeneous industrial time series (JEPA-HITS), which learns one encoder from various domains such as battery, semiconductor chemical mechanical planarization (CMP), computer numerical control (CNC) machining, and bearing while preserving dataset-specific temporal scales. We evaluate representation reuse on regression and classification problems under reduced-label downstream training, and leave-one-domain-out (LODO) pretraining. Compared with the same architecture initialized randomly, LODO pretraining reduces RMSE from 10.72 to 7.95 on battery remaining useful life (RUL), 4.34 to 3.89 on CMP, and 13.53 to 11.72 on tool wear regression, while improving bearing fault diagnosis macro-F1 from 0.547 to 0.737. Overall results suggest that unlabeled time series measurements from heterogeneous industrial processes can provide reusable representations for downstream learning with limited labeled data.
\end{abstract}
\begin{keywords}
Self-supervised, Representation learning, Predictive maintenance, Multi-task, Multi-domain
\end{keywords}
\section{Introduction}
\label{sec:intro}


Industrial sensor signals are abundant, but measurements with degradation and fault labels are scarce because they are expensive. Self-supervised learning (SSL) \cite{unsupervised_ssl}, \cite{moment} have shown that pretrained time series representations are effective even with limited labeled data. However, existing cross-domain models demonstrate transfer learning between closely related datasets. For example, \cite{ts-jepa} demonstrated transfer from FordA to FordB and \cite{faultformer} from PU bearings to CWRU bearings, leaving transfer across physically distinct industrial domains comparatively underexplored. 

Many existing approaches address this heterogeneity through fixed input-context or modality-specific preprocessing assumptions. MOMENT \cite{moment}, for example, operates on a fixed sequence length of 512 while FISHER \cite{fisher2026} relies on short-time Fourier transform (STFT) representation and targets signals with sufficiently high sampling rates. FISHER \cite{fisher2026} and UniFault \cite{UniFault} successfully demonstrate transfer across heterogeneous industrial time series data but their evaluation is primarily on rotating machineries and fault diagnosis.


Moreover, several large pretrained models \cite{das2024timesfm}, \cite{rasul2024lagllama}, \cite{ansari2024chronos} are limited to forecasting tasks. In contrast, many industrial monitoring problems require sequence-level regression tasks, such as RUL regression, average material removal rate (MRR) prediction, and tool wear prediction. The Joint-embedding predictive architecture (JEPA) \cite{jepa} provides an appealing alternative by predicting missing information in the latent space rather than the signal space. Recent TS-JEPA \cite{ts-jepa} extends latent predictive learning to time series, while TP-JEPA \cite{tp-jepa} applies JEPA to industrial prognostics. These studies motivate JEPA for temporal and industrial representation learning, but do not address joint pretraining across heterogeneous industrial processes within a shared encoder.

In this paper, we investigate whether a single JEPA encoder can learn transferable representations across physically distinct industrial domains, sensor characteristics, sequence scales, and downstream objectives. Table \ref{tab:dataset_heterogeneity} details the heterogeneity of the four industrial time series datasets. Our contributions are summarized as follows:

\begin{table*}[htbp]
\fontsize{9}{10}\selectfont
    \centering
    \caption{\fontsize{9}{10}\selectfont{Heterogeneity of industrial time series data in terms of task, sensor signals, total number of samples, median sequence length per sample, and sampling rate.}}
    \begin{tabular}{lllllll}
    \toprule
        Domain & Dataset & Task & Signals & Samples & Median len & Sampling  \\
        \midrule
        Battery & NASA \cite{nasa2007_dataset} & RUL reg. & voltage, current & 499 & $320$ & $\sim$0.1 Hz\\ 
        & &  & temperature \\
        Semiconductor & CMP \cite{cmp2016_dataset} & MRR reg. & pressure, flowrate & 2321 & $344$ & $\sim$1 Hz \\ 
        & &  & rotation rate, usage\\
        Machining & CNC \cite{phmap2025_dataset} & tool wear reg. & vibration (tri-axis), & 225 & $1.03\times10^6$ & 25.6 kHz\\ 
        & &  & sound\\
        Rotating machinery & PU \cite{paderborn2016_dataset} & fault cls. & vibration & 2239 & $256001$ & 64 kHz\\ 
        
    \bottomrule
    \end{tabular}

    \label{tab:dataset_heterogeneity}
\end{table*}

\begin{itemize}
\small
    \item We investigate joint self-supervised representation learning across physically distinct industrial domains whose sampling characteristics, sequence scales, channel dimensionalities, and downstream objectives differ substantially.
    \item We develop a heterogeneous JEPA formulation in which each domain retains its dataset-specific temporal representation while sharing a channel-independent latent encoder through multi-domain loss aggregation.
    \item We demonstrate transferable representation learning through label-scarcity evaluation, and leave-one-domain-out pretraining, in which the target industrial domain is completely unseen during SSL.
\end{itemize}

\section{Methodology}
\subsection{Preliminaries}

We consider self-supervised representation learning over a collection of \(D\), heterogeneous industrial time series datasets, $
\mathcal{D}=\{\mathcal{D}_1,\ldots,\mathcal{D}_D\},
$ where the \(d\)-th dataset is

\small{
\begin{equation}
    \mathcal{D}_d=
    \left\{
    \left(\mathbf{X}^{(d)}_i,y^{(d)}_i\right)
    \right\}_{i=1}^{N_d}
    \label{eqn:1}
\end{equation}}
where, $
\mathbf{X}^{(d)}_i
\in
\mathbb{R}^{T_i^{(d)}\times C_i^{(d)}}
$, \(T_i^{(d)}\) and \(C_i^{(d)}\) denote the sequence length and number of sensor channels, respectively, while \(y_i^{(d)}\) denotes a downstream target when labels are available. 

The objective is to learn a single encoder that captures transferable temporal structure across these heterogeneous domains without using downstream labels during pretraining. To avoid requiring a fixed
number or ordering of sensors, pretraining is performed in a channel-independent manner. Each channel of a multivariate sample is treated as an individual univariate sequence,

\small{
\begin{equation}
\mathbf{x}^{(d)}_{i,c}
=
\mathbf{X}^{(d)}_{i,:,c}
\in
\mathbb{R}^{T_i^{(d)}},
\qquad
c=1,\ldots,C_i^{(d)}.
\label{eqn:2}
\end{equation}}

The same encoder parameters \(\theta\) are shared across all channels and domains.

\subsection{Self-Supervised Architecture}
We build upon the TS-JEPA \cite{ts-jepa} framework and extend it to heterogeneous industrial time series through channel-independent raw-signal pretraining, dataset-dependent temporal patching, joint multi-domain loss aggregation, latent anti-collapse (AC) and spectral regularization. 

The JEPA-HITS architecture follows a JEPA \cite{jepa} framework comprising an online encoder $f_\theta$, an exponential moving average (EMA) target encoder $f_{\bar\theta}$, and a predictor $g_\phi$. Given visible patch indices $\mathcal V$ and masked indices $\mathcal M$, the online encoder extracts contextual representations while the target encoder provides stop-gradient latent targets $\mathbf z^t$. Unlike waveform-reconstruction objectives, the predictor estimates the representations $\hat{\mathbf z}$ of $\mathcal{M}$ directly in latent space. The primary objective uses Huber loss $H_{\delta}$ between layer-normalized $LN$ predicted and target representations, 

\small{
\begin{equation}
\mathcal{L}_{\mathrm{pred}}
=
\frac{1}{|\mathcal{M}|}
\sum_{m\in\mathcal{M}}
H_\delta
\left(
\operatorname{LN}(\hat{\mathbf z}_m)
-
\operatorname{LN}(\mathbf z^t_m)
\right).
\label{eqn:pred_loss}
\end{equation}}

To prevent representation collapse, we use VICReg-inspired variance and covariance regularization \cite{bardes2022iclr-vicreg} as anti-collapse loss term $\mathcal{L}_{AC}$. A spectral consistency term $\mathcal L_{\mathrm{spec}} = \left\|\mathcal S(\hat{\textbf{Z}})-\mathcal S(\textbf{Z}^t)\right\|_1$ additionally encourages agreement between the frequency structure of predicted and target latent trajectories,  where the magnitude is log-scaled \(\mathcal S(\mathbf Z)=\log(1+|\operatorname{rFFT}(\mathbf Z)|)\). The domain-specific objective is 

\small{
\begin{equation}
\mathcal{L}_d
=
\mathcal{L}_{\mathrm{pred},d}
+
\mathcal{L}_{\mathrm{AC},d}
+
\lambda_{spec}
\mathcal{L}_{\mathrm{spec},d}
\label{eqn:11}    
\end{equation}}








To jointly optimize heterogeneous datasets without forcing a common sequence length or channel dimensionality, we use domain-specific training buckets. Each source retains its dataset-specific temporal window, patch size, and mini-batch size, while channels are independently processed by the shared encoder. Domain losses are then aggregated, for all the domains $D=4$ in our case, before a single optimizer update.

\small{
\begin{equation}
    \mathcal{L}_{multi}=\frac{1}{D}\sum^D_{d=1}\mathcal{L}_d
\label{eqn:bucketed_loss_agg}
\end{equation}}

This framework, motivated by \cite{UMedPT}, permits joint pretraining of datasets that differ in sequence characteristics, and prediction task without padding or resampling to force all domains to a common sequence length, channel dimensionality, or sampling rate.

\begin{figure*}[htbp]
    \centering
    \includegraphics[width=0.95\linewidth]{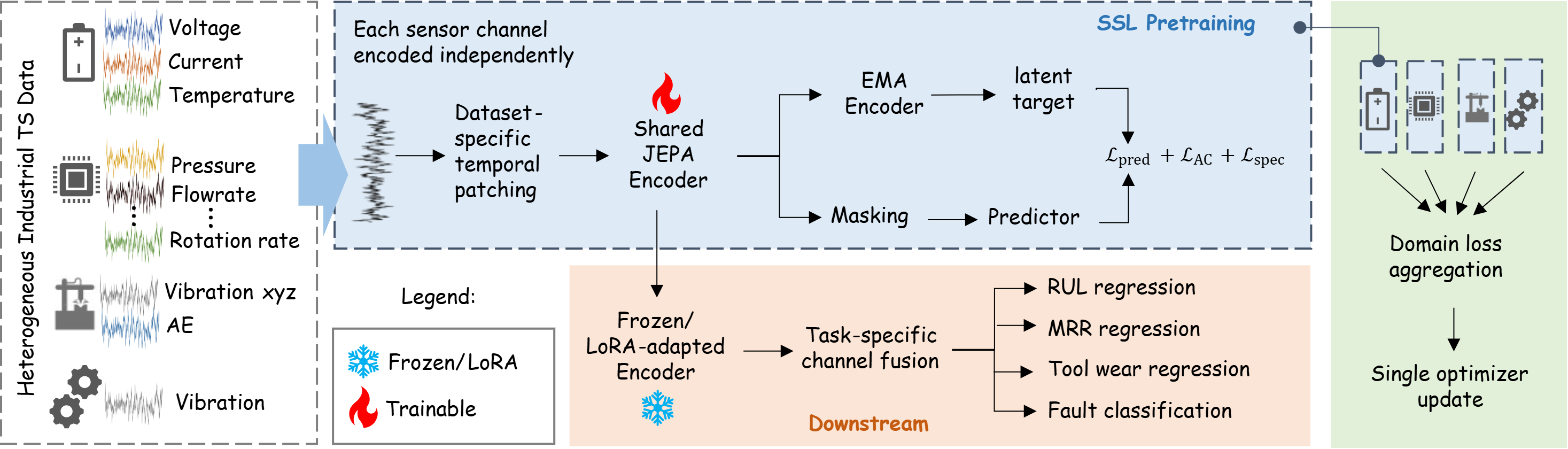}
    \caption{\small{Overview of the JEPA-HITS architecture for heterogeneous industrial time series.}}
    \label{fig:architecture}
\end{figure*}

For the downstream configuration, we use a linear head or multi-layer perceptron (MLP), frozen or low-rank adaptation (LoRA) \cite{lora} of rank four depending on validation results, yielding frozen-linear for NASA, LoRA-MLP for CNC, and frozen-MLP for both CMP and PU. Figure \ref{fig:architecture} shows the overall architecture of the JEPA-HITS framework.

\section{Experiments}

\subsection{Experimental Setup}

\textbf{Datasets}. We evaluate our proposed framework on four different industrial domains such as battery, semiconductor manufacturing, computer numerical control (CNC) machining, and rotating machinery represented by the following datasets: National Aeronautics and Space Administration (NASA) Prognostics Center of Excellence (PCoE) battery \cite{nasa2007_dataset},
Chemical Mechanical Planarization (CMP) PHM 2016 data challenge \cite{cmp2016_dataset} for material removal rate (MRR) prediction,  CNC tool wear PHM-Asia Pacific (PHM-AP) 2025 data challenge \cite{phmap2025_dataset}, and Paderborn University (PU) bearing \cite{paderborn2016_dataset}. To avoid data leakage, we perform transparent pretraining from scratch and train-test partition before preprocessing. For CMP and CNC datasets, we use the official train-test splits of the data challenge \cite{cmp2016_dataset}\cite{phmap2025_dataset} with specific chamber IDs \{4,5,6\} for CMP, while for NASA and PU datasets we use the train-test splits in \cite{nasa_sota} and \cite{paderborn_test_splits}. All self-supervised pretraining and downstream model selection use training partitions only.
\newline

\noindent\textbf{Implementation Details}.
Experiments were conducted on an NVIDIA RTX 3070 8-GB GPU computer. Experiments were run on three different seeds and the average values are reported in the results. We compare our model against fully-supervised models MiniRocket \cite{minirocket} and PatchTST \cite{Yuqietal-2023-PatchTST}, pretrained MOMENT-small \cite{moment} and FISHER-small \cite{fisher2026}, and baseline TS-JEPA \cite{ts-jepa}, which was pretrained on each dataset depending on the target domain. Pretrained models are adapted using rank-4 LoRA and an MLP head. FISHER is not evaluated for NASA and CMP to avoid artificially resampling their low-frequency $<100$ Hz signals to satisfy its STFT requirements. JEPA-HITS uses 2,000 joint multi-domain optimizer updates, learning rate \(3\times10^{-4}\), weight decay \(10^{-4}\), and initial EMA momentum \(0.99\). For NASA/CMP/CNC/PU, respectively, mask ratios are \(0.85/0.70/0.50/0.50\), window lengths are \(256/1024/4096/4096\), and \(\lambda_{\mathrm{spec}}\) values are \(0.01/1.0/0.5/0.1\). The encoder has three Transformer layers, four heads, and \(192\)-dimensional embeddings; the predictor also uses \(192\) dimensions. Patch size is 32 except for PU, where it is 1024.
\newline

\noindent\textbf{Performance Evaluation}.
For leave-one-domain-out (LODO) evaluation, the entire target domain is additionally excluded from self-supervised pretraining. The evaluation metrics are root mean squared error for regression and macro-F1, which is the average of F1 scores for all classes, for classification. The label efficiency plot uses coefficient of determination $R^2$ to visualize comparison with the baseline.
\newline


\begin{figure}[htbp]
  \centering
  \includegraphics[width=0.92\linewidth]{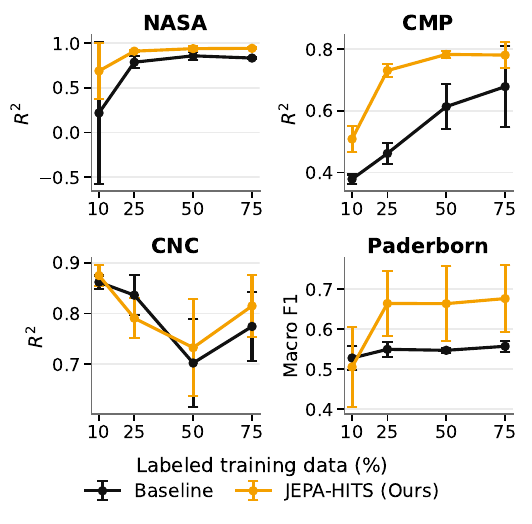}
    \caption{\small{Label efficiency of JEPA-HITS on NASA, CMP, CNC, and PU compared against the same architecture with randomly initialized encoder.}}
    \label{fig:label efficiency}
\end{figure}

\begin{table*}[!htbp]
\centering
\fontsize{9}{10}\selectfont
\caption{\fontsize{9}{10}\selectfont{Performance comparison of JEPA-HITS against different models across datasets.}}
\label{tab:results}
\begin{tabular}{lc|c|c|c|c}
\toprule
& NASA 
& CMP
& CNC
& PU
& \# Params \\

Model & (RMSE) $\downarrow$  & (RMSE) $\downarrow$  & (RMSE) $\downarrow$  & (macro-F1) $\uparrow$ & (M) $\downarrow$\\
\midrule

MiniRocket \cite{minirocket} & 29.44 $\pm$ 1.01 & 4.02 $\pm$ 0.05  & 28.97 $\pm$ 3.82 & 0.4962 $\pm$ 0.01 & --- \\
PatchTST \cite{Yuqietal-2023-PatchTST}  & 34.84 $\pm$ 4.37  & 8.39 $\pm$ 0.36  & 12.87 $\pm$ 2.51 & 0.1851 $\pm$ 0.00 & 0.135 \\
MOMENT-small \cite{moment} & 40.53 $\pm$ 5.61 & 5.34 $\pm$ 0.03  & 12.95 $\pm$ 1.10 & 0.5824 $\pm$ 0.04 & 37.9\\
FISHER-small \cite{fisher2026}  & --- & ---  & 14.54 $\pm$ 2.52 & 0.4860 $\pm$ 0.04 & 21.4 \\
TS-JEPA (adapted) \cite{ts-jepa} & 19.40 $\pm$ 1.96 & 4.09 $\pm$ 0.07 & 31.02 $\pm$ 19.84 & 0.6147 $\pm$ 0.03 & 0.363\\
Random-init JEPA-HITS & 10.72 $\pm$ 1.01 & 4.34 $\pm$ 0.26 & 13.53 $\pm$ 0.61 & 0.5466 $\pm$ 0.01 & 1.55\\
JEPA-HITS$_{LODO}$ & 7.95 $\pm$ 0.00 & 3.89 $\pm$ 0.10 & 11.72 $\pm$ 0.74 & 0.7367 $\pm$ 0.01 & 1.55 \\
JEPA-HITS & \textbf{6.65 $\pm$ 0.00} & \textbf{3.87 $\pm$ 0.04} & \textbf{10.47 $\pm$ 0.54} & \textbf{0.7793 $\pm$ 0.04} & 1.55 \\

\bottomrule
\end{tabular}
\label{tab:benchmark_results}
\end{table*}

\subsection{Results and Discussion}
Table \ref{tab:benchmark_results} shows that across all four downstream datasets, the pretrained encoder improves upon the same 1.55M-parameter architecture initialized randomly. The largest differences occur on NASA and Paderborn, while CMP shows a smaller but consistent improvement. This comparison establishes that the downstream advantage cannot be explained solely by increasing the capacity of the TS-JEPA architecture. In LODO pretraining, the representation learned exclusively from the other three domains reaches 7.95, 3.89, and 11.72 RMSE on NASA, CMP, and CNC and 0.737 macro-F1 on Paderborn after downstream adaptation. This indicates that at least part of the pretraining benefit arises from features shared across industrial systems rather than unlabeled exposure to the target data itself. We additionally evaluated learning-progress-based adaptive domain weighting, however, it did not provide consistent improvements across domains. 

\subsection{Label Efficiency}
Figure \ref{fig:label efficiency} shows that the effect of pretraining is not restricted to the fully supervised setting. The largest gains occur for NASA, CMP, and PU showing a persistent advantage with $63.70\%$, $33.79\%$ average $R^2$ and $14.85\%$ average macro-F1 improvements relative to baseline across the evaluated fractions. CNC is less consistent, with $1.40\%$ average relative $R^2$ improvement, where pretraining does not dominate the random encoder at every intermediate fraction, indicating that the usefulness of the learned prior depends on the downstream process and supervision regime.

\subsection{Ablation Study}
Table \ref{tab:ablation_loss} shows that the addition of both anti-collapse (AC) and spectral (spec) loss regularizers to JEPA-HITS yielded 38.31\%, 4.69\%, and 1.13\% reductions in RMSE for NASA, CMP, and CNC respectively, while 2.75\% improvement in macro-F1 on PU. The AC loss and spectral loss provide domain-dependent effects when applied individually, whereas their combination provides the strongest overall cross-domain configuration.

\begin{table}[t]
\fontsize{9}{10}\selectfont
\centering
\caption{\small{Ablation study of the loss components.}}
\label{tab:ablation}
\begin{tabular}{cc|cccc}
\toprule
\multicolumn{2}{c|}{Model} & \multicolumn{4}{c}{Dataset} \\
AC & spec & NASA$\downarrow$ & CMP$\downarrow$ & CNC$\downarrow$ & PU$\uparrow$ \\
\midrule
 &  & 10.78 & 4.05 & 10.59 & 0.7578 \\
& \checkmark & 18.07 & 4.01 & \textbf{10.08} & 0.6819 \\
\checkmark & & 7.35 & 3.92 & 10.55 & 0.6077 \\
\checkmark & \checkmark & \textbf{6.65} & \textbf{3.86} & 10.47 & \textbf{0.7793} \\
\bottomrule
\label{tab:ablation_loss}
\end{tabular}
\end{table}

\section{Conclusion}

We studied whether unlabeled measurements from physically distinct industrial processes can improve downstream learning when labels are scarce. A shared JEPA representation pretrained across battery, semiconductor, machining, and bearing data consistently improves a same capacity random baseline under full supervision and remains useful when each target domain is removed from pretraining. Although the magnitude of the gain varies by domain and label fraction, the results provide evidence that heterogeneous industrial sensor collections contain transferable representation structure. This suggests a practical route toward reusable industrial pretrained encoders without requiring every downstream asset to contribute unlabeled pretraining data. Our work is not evaluated on extreme cases with signal outliers, missing data, and distribution shifts. Future work will extend JEPA-HITS to additional industrial domains and downstream tasks, including forecasting, leveraging its shared encoder design to scale across heterogeneous time series data and tasks.

\vfill
\pagebreak

\clearpage
\bibliographystyle{IEEEbib}
\bibliography{strings,refs}

\clearpage
\onecolumn
\appendix

\end{document}